\documentclass{article} 
\usepackage{iclr2023_conference,times}

\usepackage[utf8]{inputenc} 
\usepackage[T1]{fontenc}    
\usepackage{hyperref}       
\usepackage{url}            
\usepackage{booktabs}       
\usepackage{amsfonts}       
\usepackage{nicefrac}       
\usepackage{microtype}      
\usepackage{xcolor}         
\usepackage{times}
\usepackage{epsfig}
\usepackage{graphicx}
\usepackage{amsmath}
\usepackage{amssymb, amsmath,amsfonts,amsthm,amssymb,mathtools,graphicx,bm}
\usepackage{multirow}
\usepackage{subfigure}
\newcommand{\bx}{\mathbf{x}}

\newcommand{\bz}{\mathbf{z}}
\newcommand{\bg}{\mathbf{g}}

\newcommand{\by}{\mathbf{y}}

\newcommand{\bh}{\mathbf{h}}
\newcommand{\bb}{\mathbf{b}}
\newcommand{\br}{\mathbf{r}}
\newcommand{\bt}{\mathbf{t}}

\usepackage{easyReview}
\usepackage{todonotes}
\usepackage{algorithmic}
\usepackage{wrapfig}
\usepackage{lipsum}
\usepackage[ruled]{algorithm2e}
\usepackage{wrapfig}

\usepackage{hyperref}
\usepackage{url}

\title{Continuous Depth Recurrent Neural Differential Equations}

\author{Srinivas Anumasa, Geetakrishnasai Gunapati  \& P.K. Srijith  \\
Department of Computer Science and Engineering\\
Indian Institute of Technology Hyderabad,India\\
}

\iclrfinalcopy 
\begin{document}

\maketitle

\begin{abstract}
Recurrent neural networks (RNNs) have brought a lot of advancements in sequence labeling tasks and sequence data. However, their effectiveness is limited  when the observations in the sequence are irregularly sampled, where the observations arrive at irregular time intervals. To address this, continuous time variants of the RNNs  were introduced based on neural  ordinary differential equations (NODE). They  learn a better representation of the data using the continuous transformation of hidden states over time, taking into account the time interval between the observations. However, they are still limited in their capability as they use the discrete transformations and a fixed discrete number of layers (depth) over an  input in the sequence to produce the output observation. We intend to address this limitation by proposing  RNNs based on  differential equations which model continuous  transformations over both  depth and time to predict an output for a given input in the sequence. Specifically, we propose continuous depth recurrent neural  differential equations (CDR-NDE) which generalizes  RNN models by continuously evolving the hidden states in both  the temporal and depth dimensions. CDR-NDE considers two separate differential equations over each of these dimensions and models the evolution in  the temporal and depth directions alternatively. We also propose the CDR-NDE-heat model based on partial differential equations which treats the computation of hidden states as solving a heat equation over time.  We demonstrate the effectiveness of the proposed models by comparing against the  state-of-the-art RNN models on  real world sequence labeling problems and data.
\end{abstract}

\section{Introduction} 
Deep learning models such as ResNets~\citep{he2016resnet} have brought a lot of advances in  many real world computer vision applications~\citep{ren17,he20,wang2019hierarchical}. They  managed to achieve a good generalization performance  by addressing the vanishing gradient problem in deep learning using skip connections. Recently, it was shown that the  transformation of hidden representations in the ResNet block is similar to the Euler numerical method~\citep{lu2018beyond,haber2017stable} for solving ordinary differential equations (ODE) with constant step size. This observation has led to the inception of new  deep learning architectures based on differential equations such as neural ODE (NODE)~\citep{node}. NODE performs continuous transformation of hidden representation by treating Resnet operations as an ODE parameterized by a neural network and solving the ODE using  numerical methods such as Euler method  and  Dopri-5~\citep{kimura2009dormand}. NODE automated the model selection (depth estimation), is parameter efficient and is robust  towards adversarial attacks than a  ResNet with similar architecture~\citep{robustode}.

Recurrent neural networks and its variants such as long short term memory (LSTM)~\citep{lstm} and gated recurrent units (GRU)~\citep{gru} were successful and effective in modeling time-series and sequence data. However, RNN models were not effective  for irregularly sampled time-series data~\citep{rubanova2019latent}, where the observations are measured at irregular intervals of time. ODE-RNN~\citep{rubanova2019latent} modeled hidden state transformations across time using a NODE, where the transformations of hidden representations depended on the time-gap between the arrivals and this led to a better representation of hidden state. This addressed the drawbacks of the RNN models which  performs a single transformation of the hidden representation at the observation times irrespective of the time interval. Such continuous recurrent models such as  GRU-ODE~\citep{de2019gru} and ODE-LSTM~\citep{lechner2020learning} were proposed to learn better representation of irregular time series data.
When  applied to the sequence data with a sequence of input-output elements along with their time of occurrences,  these models obtain the temporal evolution of hidden states using a neural ODE. At an observation time, this is  then combined with the input at that time and a discrete number of transformations is applied using a feed-forward neural network to obtain  the final hidden representation. This final hidden representation   is then used to produce  the desired output.  Though these models evolve continuously over time, they use a fixed discrete transformations over depth.

There are several real world sequence labelling problems where the sequences could be of different complexities or the input elements in the sequence could be of different complexities. For instance, consider the problem of social media post classification where different posts arrive at irregular time intervals. The  posts could have varying characteristics with some posts containing only text while some contains both text and image. It would be beneficial to have a recurrent neural network model which would consider the complexities of the input in a sequence by having a varying number of transformations for different inputs. In this work, we propose continuous depth recurrent neural differential equation (CDR-NDE) models which generalize the recurrent NODE models to have continuous transformation over depth in addition to the time. Continuous depth allows flexibility in modeling sequence data, with different depths over the elements in the sequence as well as different sequences. Combining this with the continuous time transformation as in recurrent neural ODE allows greater modeling capability for irregularly sampled sequence data.

The proposed continuous depth recurrent neural  differential equations (CDR-NDE)  model the evolution of the hidden states simultaneously in both the temporal and depth dimensions using differential equations.  Continuous transformation of hidden states  is modeled as a differential equation with two independent variables, one in the temporal and the other in the depth direction.  We also aim to model the  evolution of the hidden states using a partial differential equation (PDE)  based on the 1D-heat equation, leading to the CDR-NDE-heat model. Heat equation is  a second order partial differential equation, which models the flow of heat across the rod over time. The proposed CDR-NDE-heat model considers the transformation of hidden states across depth and time using a non-homogeneous heat equation. An advantage is that it is capable of considering the information from the future along with the past in sequence labeling tasks.  We exploit the structure in the CDR-NDE-heat model and PDE solvers  to develop an efficient way to obtain the hidden states where  all the hidden states at a particular depth can be computed simultaneously.   We evaluate the performance of our proposed models  on real-world datasets such as person activity recognition~\citep{asuncion2007uci}, Walker2d kinematic simulation data~\citep{lechner2020learning}   and stance classification of social media posts~\citep{RumourEval_2019_dataset}. Through experiments, we show that the proposed continuous depth recurrent neural  differential equation models outperformed the state-of-the-art recurrent neural networks in all these tasks.   

\section{Related Work} 
RNN models such as LSTM~\citep{lstm} and GRU~\citep{gru} are the primary choice to fit high-dimensional time-series and sequence data. For irregular time-series data, traditional LSTM and GRU models  are less effective as they do not consider the varying inter-arrival times. To address the problem of fitting irregular time-series data, the standard approach is the augmented-LSTM which augments the elapsed time with input data.  In GRU-D~\citep{grud} and RNN-Decay~\citep{rubanova2019latent}, the computed hidden state is the  hidden state multiplied by a decay term  proportional to the elapsed time.  In other variants such as CT-GRU~\citep{ctgru},CT-RNN~\citep{ctrnn}, ODE-RNN~\citep{rubanova2019latent}, GRU-ODE~\citep{de2019gru}, ODE-LSTM~\citep{lechner2020learning} and Jump-CNF~\citep{spatio-temp}, the hidden state is computed as a continuous transformation of intermediate hidden states. 
 CT-LSTM~\citep{ctlstm} combines both LSTM and continuous time neural Hawkes process to model continuous transformation of hidden states. Two alternative states are computed at each time-step and the final state is an interpolated value of these hidden states, where the interpolation depends on the elapsed time. Phased-LSTM~\citep{neil2016phased} models irregularly sampled data using  an additional time gate. The updates to the cell state and hidden state only happen when the time gate is open. 
 This time gate allows for the updates to happen at irregular intervals. Phased LSTM reduces the memory decay as the updates only happen in a small time when the time gate is open. ODE-RNN~\citep{rubanova2019latent} used neural ordinary differential equations over time to model the evolution of the hidden states. The next hidden state is obtained as a solution to a NODE, and depends on the time interval between two consecutive observations. GRU-ODE~\citep{de2019gru} derived a NODE over time and hidden states using the GRU operations and consequently could avoid the vanishing gradient problem in  ODE-RNN. Similarly, ODE-LSTM~\citep{lechner2020learning} addressed the vanishing gradient problem in  ODE-RNN by considering the LSTM cell and memory cell while  the output state is modeled using a neural ODE to account for irregular  observations. However, all these models only considered continuous evolution of hidden states over the temporal dimension.  In our work, we aim to develop models which consider continuous evolution of the hidden states over depth as well as temporal dimensions. 

Recently, there are some works which used  deep neural networks to solve the  partial differential equations (PDE) (also known as neural PDEs or physics informed neural networks) ~\citep{zubov2021neuralpde,brandstetter2021message,Yihaoneuralpde}.
~\citep{Yihaoneuralpde} showed that LSTM based RNNs can efficiently find the solutions to multidimensional PDEs without knowing the specific form of PDE. On the other hand, very few works used PDEs to model DNN architectures for solving problems from any domain. \citep{RuthottoDNNpde} used PDE to design Resnet architectures and  convolutional neural networks (CNNs) such as   Parabolic CNN and  hyperbolic CNN by changing the ODE update dynamics to different PDE update dynamics.  For instance, hyperbolic CNN can be obtained with second order dynamics. They showed that even the PDE CNNs with modest architectures achieve similar performance to the larger networks with considerably large numbers of parameters. 
Unlike the prior works combining neural networks and PDEs, our aim is to solve sequence labeling problems by developing flexible RNN based architectures considering the PDE based models and solutions. 

 
\section{Background}

\subsection{Problem Definition}
We consider the sequence labeling problem with a sequence length of $K$, and denote the input-output pairs in the sequence as $\{\bx_t, \by_t\}_{t=1}^K$ and  the elements in the sequence are irregularly sampled at observation times $\bt \in \mathbb{R^+}^K$. We assume  the input element in the sequence to be  $D$ dimensional, $\bx_t \in \mathcal{R}^D$ and the  corresponding output $\by_t$ depends on the problem, discrete  if it is classification or continuous if it is  regression. The aim is to learn a model $f(\cdot, \theta)$ which could predict the output $\by_t$ considering the input $\bx_t$, and dependence on other elements in the sequence. 

\subsection{Gated Recurrent Unit}
Recurrent neural  networks (RNNs) are well suited to model the sequence data. They make use of the recurrent connections to remember the information until the previous time step, and combine it with the current input to predict the output. Standard RNNs suffer from the vanishing gradient problem due to which it forgets long term dependencies among the sequence elements. This was overcome with the help of long short term memory (LSTM)~\citep{lstm} and gated recurrent units (GRUs)~\citep{gru}. In our work we consider the basic RNN  block to be a  GRU. In GRU, computation of  hidden state and output at any time step $t$ involves the following transformations,
\begin{equation}
    \begin{aligned}
    \br_t &= \sigma(W_r\bx_t + U_r\bh_{t-1} + \bb_r)  , \quad 
    \bz_t = \sigma(W_z\bx_t + U_z\bh_{t-1} + \bb_z) \\
    \bg_t &= \tanh(W_h\bx_t + U_h(\br_t \odot \bh_{t-1}) + \bb_h) 
    \label{eq:gru}
    \end{aligned}
\end{equation}

Where $\br_t$,$\bz_t$,$\bg_t$ are the reset gate, update gate and update vector respectively for the GRU. The  hidden state $\bh_t$ in GRU is given by,
\begin{equation}
    \bh_t = \bz_t \odot \bh_{t-1}  + (1-\bz_t)\odot\bg_t
    \label{eq:gru1}
\end{equation}
As we can see, GRUs and RNNs in general do not consider the exact times or time interval between the observations. The same operations are applied irrespective of the  time gap between observations. This can limit the capability of these models for irregularly sampled time series data.  

GRUs can be extended to consider the irregularity in the time series data by developing a continuous GRU variant. A continuous GRU, GRU-ODE~\citep{de2019gru}, can be obtained by adding and  subtracting the hidden state on both sides of \eqref{eq:gru1}. The  computation of hidden states then becomes equivalent to solving an ODE given in \eqref{eq:gru-ode}. 
\begin{equation}\hspace{-3mm}
    \begin{aligned}
    \Delta\bh_t &= \bh_t - \bh_{t-1}= \bz_t \odot \bh_{t-1}  + (1-\bz_t)\odot\bg_t - \bh_{t-1} 
    \implies \frac{d\bh_t}{dt} &= (1-\bz_{t}) \odot (\bg_{t} - \bh_{t})
    \label{eq:gru-ode}
    \end{aligned}
\end{equation}

\subsection{Recurrent Neural Ordinary Differential Equations}
In ODE-RNN~\citep{odernn}, ODE-LSTM~\citep{lechner2020learning}, and ODE-GRU~\citep{de2019gru},  the hidden state $\bh_t$ holds the summary of past observations and evolves continuously between the observations considering the time interval. When there is a new observation, the hidden state $\bh_t$ changes abruptly~\citep{spatio-temp} to consider it. Given initial state $\bh_0$, let the function $f_h()$ models the continuous transformation of hidden state and function $g_h()$ model instantaneous change in the hidden state at the new observation. The prediction $y'_t$ is computed by using a function $o_h()$, which is then used to compute the loss (cross entropy loss for classification). Both the functions $g_h()$ and $o_h()$ are typically standard feed-forward neural networks with a discrete number of layers.  In the case of GRU-ODE, the function $f_h()$ takes the form given in the right hand side of \eqref{eq:gru-ode}. In general, the recurrent NODE models can be represented using the following system.
\begin{equation}
    \frac{d\bh_t}{dt} = f_h(h_t) , \quad \lim_{\epsilon\to0} \bh_{t+\epsilon} = g_h(\bh_t,\bx_t)  , \quad  y'_t = o_h(\bh_{t+\epsilon})
\end{equation}

\section{Continuous Depth Recurrent Neural Differential Equations}
We propose continuous depth recurrent neural differential equations (CDR-NDE) aiming to overcome the drawbacks of the recurrent NODEs in modeling the sequence data. As already discussed, recurrent NODE models bring abrupt changes in  the hidden state at the observation times using the standard neural network transformation $g_h()$, when it considers the input $\bx_t$ at time $t$. We aim to develop RNN models capable of continuous transformations over depth in addition to the temporal dimension. Such models will  help in processing inputs with varying complexities,  aid in model selection (in choosing the number of layers in a RNN model) and reduce the number of parameters (as every layer shares  the same parameters as in NODE). Moreover, we hypothesize that such continuous transformation over depth will also aid in learning better hidden representations as it is not limited by a predefined number of layers as in standard RNNs. 
We propose two models, CDR-NDE and CDR-NDE-heat model. Both the models generalize the GRU-ODE model by continuously evolving in both the temporal  and depth directions. The CDR-NDE-heat model is formulated based on a partial differential equation (1-dimensional heat equation) and enables faster computation of hidden states even when using an adaptive step numerical method like Dopri5.

\begin{figure}[t]
  \begin{center}
    \includegraphics[scale=0.4]{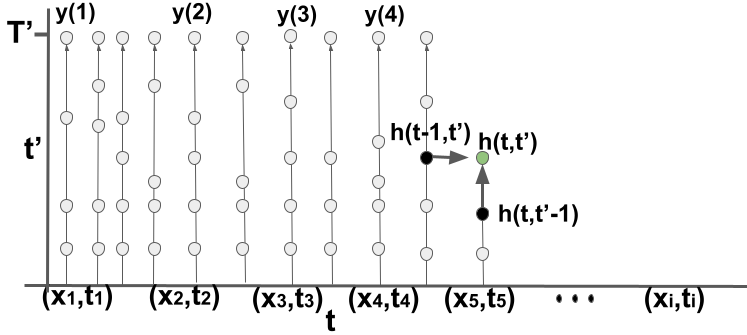}
  \end{center}
  \caption{ Shows the computation of hidd   en states using the CDR-NDE model. As we can see for different observations, the number of intermediate hidden states in the evolution along the depth is different. The index $(t-1,t')$ points to the hidden state of immediate left vertical evolution. The index $(t,t'-1)$ points the hidden state just below the current vertical evolution. }
  \label{fig:deep-gru-ode}
  \end{figure}

The  hidden states of CDR-NDE  evolves in depth direction (denoted by vertical axis $t'$ in Figure~\ref{fig:deep-gru-ode}) while also evolving in the  temporal direction (denoted by horizontal axis  $t$ in  Figure~\ref{fig:deep-gru-ode}).  As the hidden state evolves in both directions,  we need a tuple $(t,t')$ to uniquely identify any hidden state and we represent the hidden state as $\bh_{(t,t')}$. 
As shown in Figure~\ref{fig:deep-gru-ode}, during  the computation of a hidden state  $\bh_{(t,t')}$, it requires hidden states that are immediately to the left in the temporal dimension $\bh_{(t-1,t')}$  and below in the depth direction $\bh_{(t, t'-1)}$. The evolution of the  hidden state  $\bh_{(t,t')}$ in the CDR-NDE is governed by the following differential equations. 
\begin{eqnarray}
   \frac{\partial \bh_{(t,t')}}{\partial t} &=& f_h(\bh_{(t,t')},\bh_{(t,t'-1)})  , \quad
   \frac{\partial \bh_{(t,t')}}{\partial t'} = g_h(\bh_{(t,t')},\bh_{(t-1,t')})\\
   \by_i &=& o_h(\bh_{(t_i,T')})  , \quad
    \bh_{(t_i,0)} = \bx_i \quad \forall i=1,\ldots,K 
\end{eqnarray}
where $T'$ is the maximum depth. We observe that the changes in the hidden state in the horizontal (time) direction depends on the hidden states at a depth below while changes in the hidden states in the vertical (depth) direction depends on the hidden states at the previous time.  The derivation and  exact expression used to define the functions $f_h()$ and $g_h()$ are obtained as follows. We consider the evolution in the horizontal direction to follow the GRU-ODE model but with an added skip-connection in the vertical direction. Though $f_h()$ can be any function as in ODE-RNN, we followed GRU-ODE to avoid vanishing gradient problems in the temporal direction~\citep{de2019gru}. In a discrete setup, the expression used to compute the hidden state $\bh_{(t,t')}$ after adding the skip connection in the vertical direction can be written as 
\begin{equation}
    \bh_{(t,t')}  = \bz_{(t,t')} \odot \bh_{(t-1,t')}  + (1-\bz_{(t,t')})\odot\bg_{(t,t')} +  \bh_{(t,t'-1)}
    \label{eq:gru-ode-deep}
\end{equation}
By subtracting $\bh_{(t-1,t')}$ on both sides, we can obtain  the difference equation as 
\begin{equation}
    \begin{aligned} 
    \bh_{(t,t')} - \bh_{(t-1,t')} &= \bz_{(t,t')} \odot \bh_{(t-1,t')}  + (1-\bz_{(t,t')})\odot\bg_{(t,t')} +\bh_{(t,t'-1)} - \bh_{(t-1,t')} 
    \end{aligned}
\end{equation}
Consequently, the  differential equation governing the flow in the temporal (horizontal) direction is 
\begin{equation}
    \begin{aligned} 
    \frac{\partial \bh_{(t,t')}}{\partial t} &= \bz_{(t,t')} \odot \bh_{(t,t')}  + (1-\bz_{(t,t')})\odot\bg_{(t,t')} +\bh_{(t,t'-1)} - \bh_{(t,t')}
    \end{aligned}
    \label{eq:cdrnde1}
\end{equation}
where 
\begin{equation}
    \begin{aligned}
    \bz_{(t,t')} &= \sigma(W_z\bh_{(t,t'-1)} + U_z\bh_{(t,t')} + \bb_z)\nonumber, \quad
    \bg_{(t,t')} = \tanh(W_h\bh_{(t,t'-1)} + U_h(\br_{(t,t')} \odot \bh_{t,t'}) + \bb_h) \\
    \br_{(t,t')} &= \sigma(W_r\bh_{(t,t'-1)} + U_r\bh_{(t,t')} + \bb_r) \nonumber 
    \end{aligned}
\end{equation}
To derive the differential equation in the  depth (vertical) direction $t'$, Equation \ref{eq:gru-ode-deep} can be written as a difference equation by carrying the term $\bh_{(t,t'-1)}$ to the left hand side. 
\begin{equation}
    \begin{aligned}
    \bh_{(t,t')} - \bh_{(t,t'-1)} &= \bz_{(t,t')} \odot \bh_{(t-1,t')}  + (1-\bz_{(t,t')})\odot\bg_{(t,t')} \\
    \end{aligned}
\end{equation}
Consequently, the  differential equation governing the flow in the depth (vertical) direction is defined below and we can observe that it depends on the  hidden states at the previous time.
\begin{equation}
    \begin{aligned}
   \frac{\partial \bh_{(t,t')}}{\partial t'} &= \bz'_{(t,t')} \odot \bh_{(t-1,t')}  + (1-\bz'_{(t,t')})\odot\bg'_{(t,t')} 
    \end{aligned}
    \label{eq:cdrnde2}
\end{equation}
where
\begin{equation}
    \begin{aligned}
    \bz'_{(t,t')} &= \sigma(W_z\bh_{(t,t')} + U_z\bh_{(t-1,t')} + \bb_z)\nonumber, \quad 
    \bg'_{(t,t')} = \tanh(W_h\bh_{(t,t')} + U_h(\br'_{(t,t')} \odot \bh_{(t-1,t')}) + \bb_h) \\
        \br'_{(t,t')} &= \sigma(W_r\bh_{(t,t')} + U_r\bh_{(t-1,t')} + \bb_r) \nonumber  
    \end{aligned}
\end{equation}
We solve the differential equations \eqref{eq:cdrnde1} and \eqref{eq:cdrnde2} in two stages.  In the first stage, CDR-NDE is solved in the horizontal direction until time $t_K$ for $t'=0$ following \eqref{eq:cdrnde1} and can be solved using differential equation solvers such as Euler method or Dopri5. 
In the second stage, for every  time step evaluated  on the $t$-axis during the first stage, hidden states are allowed to evolve in the vertical direction, i.e.  along the $t'$-axis.  Evolution in vertical direction is done until time $t'= T'$ and can be solved  using  solvers such as Euler or Dopri5. We can observe that during this evolution, CDR-NDE model considers  $\bh_{(t-1,t')}$ in computing $\bh_{(t,t')} $ for any time t and depth t' taking into account the dependencies in the sequence. Hence, computation of the hidden state $\bh_{(t,t')}$ needs access to the hidden state  $\bh_{(t-1,t')}$  and this requires performing an additional interpolation on the hidden states evaluated at time $t-1$ in the case of adaptive solvers. 
\subsection{ CDR-NDE based on Heat Equation } 
\begin{figure*}[t]
\begin{center}     
\subfigure[At $t'=0$ sec]{\includegraphics[scale=0.18]{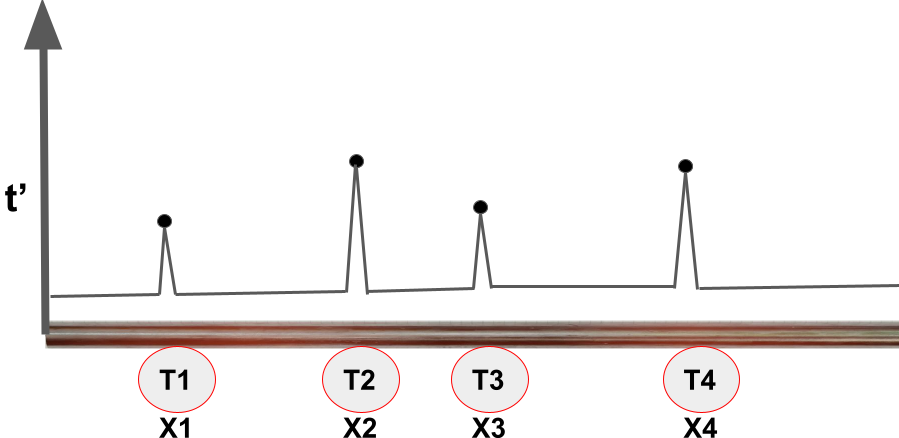}}
\subfigure[At $t'=p$ sec]{\includegraphics[scale=0.18]{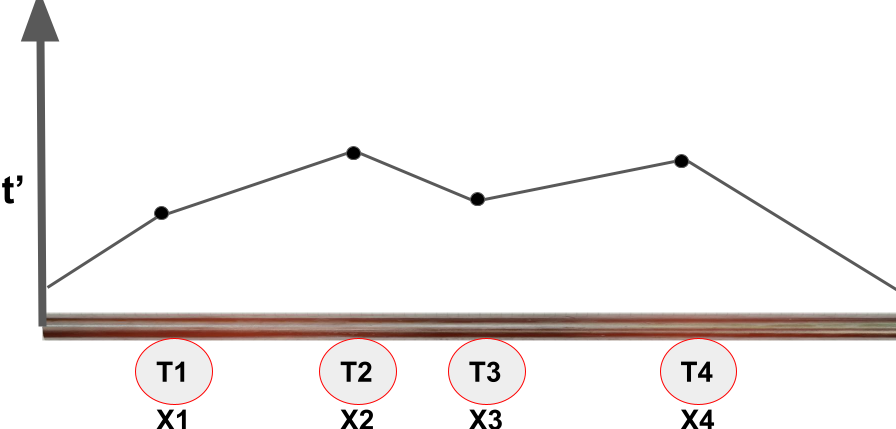}}
\subfigure[At $t'>p$ sec]{\includegraphics[scale=0.18]{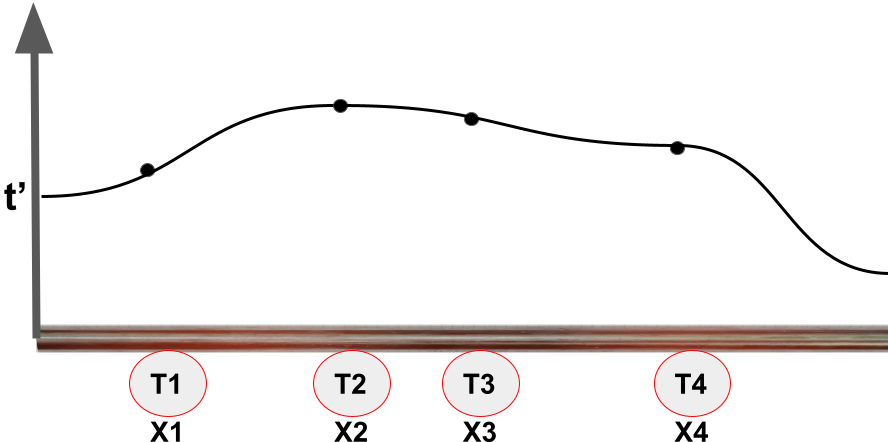}}
\end{center}
\caption{Evolution of temperature across time.(a) shows the initial state(temperature) of the rod, heat is applied externally at 4 different points. Over time, heat diffuses from hot region to cold region, (b) shows the state of the rod, after $p$ seconds. (c) Over time, the change in temperature comes to an equilibrium state, no change of temperature over time. }
\label{fig:heat}
\end{figure*}
We propose another CDR-NDE model inspired by the partial differential equations and in particular the 1D-heat diffusion equation~\citep{cannon1984one}. Heat equation represents the evolution of heat over time in a rod. Consider a rod of length $L$ which is at room temperature. Along the length of the rod, at different points, heat is applied externally into the rod. The temperatures applied at different points can be different. Figure \ref{fig:heat}(a) provides a visualization of the initial state(at $t'=0$), where the rod is at room temperature. At four different points, heat is applied externally with different temperature values. Heat starts to flow from hotter regions to colder regions and the points which are initially at room temperature become hotter. In a finite amount of time, it reaches a state where the change in temperature at any point in time is zero, which is  called an equilibrium state. Figure\ref{fig:heat}(b) visualizes the intermediate state of temperatures across the rod after $p$ seconds. Figure\ref{fig:heat}(c) visualizes the equilibrium state where the change in temperature across the rod is smooth.

We can observe that the evolution of  temperature in a rod can be seen as equivalent to the evolution of hidden states and applying temperature to the road  can be  considered equivalent to providing input data in the sequence. The hidden states associated with input data smoothly change over the depth dimension $t'$ reaching an equilibrium state, and finally leading to the output elements in the sequence.  The heat equation motivates us to construct a model capable of capturing interactions among different elements in the sequence, providing smooth hidden transformations with variable depth. We hypothesize that such models will be able to learn better representations depending on the input and improve the generalization performance.

The process of heat diffusion can be represented by a 1D heat equation~\citep{cannon1984one} which is a homogeneous second order partial differential equation,
     $\frac{\partial u(t',l)}{\partial t'} = C* \frac{\partial^{2} u(t',l)}{\partial l^{2}}$,
where $u(t',l)$ is the temperature at point $l$ on rod at time $t'$ and $C$ is a constant(diffusivity). The proposed CDR-NDE-heat model is based on the non-homogeneous heat equation, a variant of the homogeneous 1D heat equation. The temperature applied at a location $l_i$ in the rod is equivalent to the datapoint $\bx_i$ at time $t_i$. As the temperature injected at a point affects the  temperature around the rod neighborhood, the hidden states are affected by the observed data points in the neighborhood.  The hidden state then evolves over the depth variable $t'$ and reaches a steady state following  the heat diffusion model. The second order derivative with respect to location $l$  in the heat equation (equivalently over time $t$ for sequence labeling problems) allows one to consider the effect of neighbors around a point. For sequence modeling, this allows one to learn a  better representation  by considering  past and future hidden states across time $t$, similar to bi-directional RNNs considering past and future information.

The proposed model considers a non-homogeneous heat equation model which allows a better representation of hidden state during evolution by considering additional information on the interaction between the hidden states. In our case, we choose GRUcell which holds a summary of the past observations to capture the interaction.   The differential equation governing the evolution of the  proposed CDR-NDE-heat model is defined  as follows, 
\begin{equation}
     \frac{\partial \bh_{(t,t')}}{\partial t'} - \frac{\partial^{2} \bh_{(t,t')}}{\partial t^{2}} = f(\bh_{(t,t'-1)},\bh_{(t-1,t')}) 
     \label{eq:gru-heat}
\end{equation}
where $f(\bh_{(t,t'-1)},\bh_{(t-1,t')})$ is the  GRUCell  operation, i.e. $f(\bh_{(t,t'-1)},\bh_{(t-1,t')}) = \bz_{(t,t')} \odot \bh_{(t-1,t')}  + (1-\bz_{(t,t')})\odot\bg_{(t,t')}$. The evolution of hidden state as shown in Equation\ref{eq:gru-heat} corresponds to a  non-homogeneous heat equation~\citep{trong2005nonhomogeneous} with GRUCell capturing the interactions.

\begin{wrapfigure}{r}{0.5\textwidth}
  \begin{center}
    \includegraphics[scale=0.3]{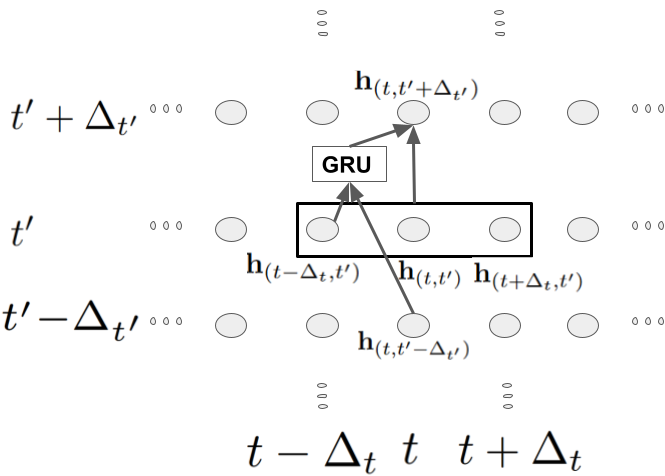}
  \end{center}
  \caption{Pictorial representation of computing a hidden state $\bh_{(t,t'+\Delta_{t'})}$ as shown in Equation \ref{eq:fdm}. The new hidden state depends on the already computed hidden states in the lower layers.}
  \label{fig:gru-heat}
\end{wrapfigure}

The heat  equation can be solved numerically using methods like finite-difference method(FDM)~\citep{recktenwald2004finite} and method of lines(MoL)~\citep{schiesser2012numerical}. We can get a better insights  on the behaviour of the proposed CDR-NDE-heat model by writing the updates using the finite difference method.  Using FDM, the hidden state is computed as follows,
\begin{multline}
    \frac{\bh_{(t,t'+\Delta_{t'})} - \bh_{(t,t')}} {\Delta_{t'}} - \frac{\bh_{(t-\Delta_{t},t')} -2\bh_{(t,t')} + \bh_{(t +\Delta_{t},t')}} {\Delta_{t}^2} =  f(\bh_{(t,t'-\Delta_{t'})},\bh_{(t-\Delta_{t},t')}) \\
 \implies  \bh_{(t,t'+\Delta_{t'})} = \frac{\Delta_t'}{\Delta_{t}^2}[\bh_{(t-\Delta_{t},t')} -2\bh_{(t,t')} + \bh_{(t +\Delta_{t},t')}] + \Delta_{t'}[f(\bh_{(t,t'-\Delta_{t'})},\bh_{(t-\Delta_{t},t')}) ] + \bh_{(t,t')}
   \label{eq:fdm}
\end{multline}
 FDM divides the space of $(t,t')$ into finite grids as shown in Figure \ref{fig:gru-heat}.
 To compute the hidden state at a depth $t' + \Delta_{t'}$, it utilizes the hidden states computed for itself and its immediate neighbors at the previous depth ($\bh_{(t-\Delta_{t},t')}, \bh_{(t,t')},  \bh_{(t +\Delta_{t},t')}$). This helps to capture dependence among the neighboring inputs during evolution. A drawback of directly using FDM in solving the proposed CDR-NDE-heat model is that it is a slow process. It doesn't exploit the GPU power as the computations are happening in  a sequential order. 
 
 For the proposed model, from  the formulation to compute next hidden state in Equation \ref{eq:fdm} and  Figure \ref{fig:gru-heat}, we can observe that the  hidden states at $t'+\Delta_{t'}$ only depends on the hidden states computed below $t'+\Delta_{t'}$. Hence,  all the hidden states $t'+\Delta_{t'}$ can be computed simultaneously once we have hidden states at time $t'$. The  numerical techniques based on Method of lines (MoL)~\citep{schiesser2012numerical} is a good choice for such a scenario. MoL method typically discretizes and computes function values in one dimension, and  then jointly evolves in the remaining dimension to compute all the function values. In our approach, we first compute hidden states along the $t$-axis and then compute the hidden states along the $t'$-axis by posing  as a solution to the system of differential equations. The evolution of the hidden states along the  $t'$-axis is defined  by the ordinary differential equation (\ref{eq:mol}), which is derived from Equation \ref{eq:fdm}. 
 \begin{equation}
\frac{\partial \bh_{(t,t')}}{\partial t'} = g_\theta(\bh_{(t-\Delta_{t},t')}, \bh_{(t,t')}, \bh_{(t +\Delta_{t},t')}) =  \frac{\bh_{(t-\Delta_{t},t')} -2\bh_{(t,t')} + \bh_{(t +\Delta_{t},t')}} {\Delta_{t}^2} +  f(\bh_{(t,t')},\bh_{(t-\Delta_{t},t')})
 \label{eq:mol}
 \end{equation}
 \begin{equation}
\bh_{(:,T')} = \text{ODESOLVE}(g_\theta,initial\_state = \bh_{(:,0)},start\_time = 0,end\_time=T') \nonumber
\label{eq:solve}
 \end{equation}
 The initial hidden states  at $t'=0$ ($\bh_{(:,0)}$) are computed  by solving an ODE along the  $t$-axis
\begin{equation}
    \frac{d\bh_{(t,0)}}{ dt} = (1-\bz_{(t,0)}) \odot (\bg_{(t,0)} - \bh_{(t,0)})
\end{equation}
 One can select any numerical method for solving the system of ODEs. In the experiments, we evaluate the performance of the CDR-NDE-heat model  using both   Euler (CDR-NDE-heat(Euler)) and  Dopri5 (CDR-NDE-heat(Dopri5)) methods.  We can observe that CDR-NDE-heat model considers  $\bh_{(t + \Delta_{t},t')}$ in addition to $\bh_{(t-\Delta_{t},t')}$ in computing $\bh_{(t,t')} $ for any time t and depth t',  taking into account more dependencies in the sequence.


After computing the hidden states at depth $T'$, predictions are made using a fully connected neural network, i.e. $\by_i = o_h(\bh_{(t_i,T')})$. This is then used to compute loss - cross-entropy for classification and root mean square error for regression problems. The parameters of the CDR-NDE models, i.e. weight parameters of the GRUCell, are learnt by minimizing the loss computed over all the observations in the sequence and over all the sequences. The computed loss is backpropagated using either adjoint method~\citep{node} or automatic differentiation to update the model parameters. 

 \begin{table}
    \centering
    \caption{ODE solvers used for different RNODE models. For the CDR-NDE-Heat model using Dopri5, the absolute and relative tolerance values are $1e^{-3}$ and $1e^{-3}$ respectively.}
    \begin{tabular}{|c|c|c|}
        \hline
            Model & ODE-Solver & Time-step Ratio \\ 
        \hline
            CT-RNN & 4-th order Runge-Kutta & 1/3 \\
            ODE-RNN & 4-th order Runge-Kutta & 1/3 \\
            GRU-ODE & Explicit Euler & 1/4 \\
            ODE-LSTM & Explicit Euler & 1/4 \\
            CDR-NDE & Explicit Euler & 1/2 \\
            CDR-NDE-heat(Euler) & Explicit Euler & 1/2 \\
            CDR-NDE-heat(Dopri5) & Dopri5 & - \\
         \hline
    \end{tabular}
    \label{tab:solvers}
\end{table}
\section{Experiments}
To evaluate the performance of the proposed models, we conduct experiments on irregular time series datasets like person activity recognition~\citep{asuncion2007uci}, walker2d kinematic simulation~\citep{lechner2020learning} and stance classification of social media posts~\citep{RumourEval_2019_dataset}. We compare our proposed models against RNN models which are designed to tackle the irregularly sampled time-series data.
\begin{wraptable}{r}{6.3cm}
    \centering
    \caption{Hyperparameter Details}
    \begin{tabular}{|c|c|}
        \hline
            Parameter & Value  \\ 
        \hline
           Hidden state Dimension & 64 \\
            Minibatch size & 256 \\
            Optimizer & RMSprop \\
            Learning rate & $5e^{-3}$ \\
            Training epochs & 200 \\
        \hline
    \end{tabular}
    \label{tab:my_label}
\end{wraptable} 
The experimental setup such as  the numerical method, hidden state dimension and other hyperparameters is the same as in~\citep{lechner2020learning} and is provided in Table~\ref{tab:my_label}. Table \ref{tab:solvers} provides the choice of numerical methods for each model. The proposed models CDR-NDE and CDR-NDE-heat(Euler) used the Euler method with the number of steps as 2.
CDR-NDE-heat(Dopri5) used Dopri5  with the absolute and relative tolerance set to $1e^{-3}$. Scheduled learning rate decay is used with decay parameter $\gamma$ = 0.1, scheduled at epoch 100. The models are trained on Nvidia Tesla V-100 32GB GPU. 

\subsection{Baselines}
We compared our proposed models against RNN models which are designed to address the problem of fitting irregular time-series data such as
GRU-ODE~\citep{de2019gru}, CT-GRU~\citep{ctgru},CT-RNN~\citep{ctrnn},GRUD~\citep{grud},Phased-LSTM~\citep{neil2016phased},ODE-LSTM~\citep{lechner2020learning}, bidirectional-RNN~\citep{schuster1997bidirectional}, RNN decay~\citep{rubanova2019latent},Hawk-LSTM~\citep{hawk}, Augmented LSTM~\citep{lechner2020learning}, and ODE-RNN~\citep{rubanova2019latent}.

\begin{table}
    \centering
 \caption{Column 2, shows the test accuracy (mean $\pm$
std) of all the models trained on the dataset \textbf{ Person
activity recognition}. Column 3 shows the test data Mean-square error (mean $\pm$
std) of all the models trained on the datasets \textbf{Walker2d Kinematic}. For both the dataset, every model is trained for 5 times with 5 different seeds. }
\begin{tabular}{|c|c|c|}
 \hline
 \textbf{Model} & \textbf{Person Activity Recognition} & \textbf{Walker2d Kinematic} \\ 
 & Test-Accuracy & Mean-Square Error(Test-data) \\
 \hline
 Augmented-LSTM & 83.78 $\pm$ 0.41 & 1.09 $\pm$ 0.01 \\
 CT-RNN & 82.32 $\pm$ 0.83 & 1.25 $\pm$ 0.03\\
 ODE-RNN & 75.03 $\pm$ 1.87 & 1.88 $\pm$0.05\\
 ODE-LSTM & 83.77 $\pm$	0.58 & 0.91 $\pm$ 0.02\\
 CT-GRU & 83.93$\pm$ 0.86 & 1.22 $\pm$ 0.01\\
 RNN-Decay & 78.74	$\pm$ 3.65 & 1.44 $\pm$ 0.01\\
 Bidirectional-RNN & 82.86 $\pm$	1.17 & 1.09 $\pm$0.01 \\
 GRU-D & 82.52	$\pm$ 0.86 & 1.14 $\pm$ 0.01\\
 Phased-LSTM & 83.34 $\pm$	0.59 & 1.10 $\pm$0.01\\
 GRU-ODE & 82.80	$\pm$ 0.61 & 1.08 $\pm$ 0.01\\
 CT-LSTM & 83.42 $\pm$	0.69 & 1.03 $\pm$0.02\\
 \hline 
  CDR-NDE & \textbf{87.54 $\pm$ 0.34} & 0.97 $\pm$ 0.04 \\
 CDR-NDE-heat (Euler) & \textbf{88.24} $\pm$ 0.31 & 0.54 $\pm$ 0.01\\
CDR-NDE-heat (Dopri5) & \textbf{88.60 $\pm$ 0.26} & \textbf{0.49 $\pm$ 0.01}\\
 \hline
\end{tabular}
    \label{tab:results}
\end{table}
\subsection{ Person activity recognition with irregularly sampled time-series}
Dataset contains sensor data from 4 from different sensors(1 for each ankle, 1 chest and 1 belt) attached to 5 different people, performing 5 sequences of activities. The task is to classify the activity based on the sensor data.
There are 11 different activities which are reduced to 7 as suggested in ~\citep{rubanova2019latent}. Dataset is transformed such that each step in the recording contains 7 values (4 of which determine the sensor that is producing data and the other 3 are sensor data). Each recording is split into overlapping intervals of 32 (with overlap of 16) and all the sequences are combined into one dataset. Out of the total sequences 7769 used for training, 1942 used for testing. 

In Table \ref{tab:results}, column 2 shows the performance of all the models trained on a person-activity dataset in terms of test accuracy. Our proposed models CDR-NDE and CDR-NDE-heat perform better than all  other baseline models. It shows that considering the continuous transformation along both the directions results in a model with better representation capability and generalization performance. We also observe that the more flexible CDR-NDE-heat model using the adaptive Dopri5 solver gives the best performance in the person-activity dataset.  To verify the flexibility of the model and the  requirement of  different depth for different sequences,  we computed the number of function evaluations involved while evolving the hidden states over depth in the CDR-NDE-heat(Dopri5) model. We found that the number of function evaluations fall in the range of 26 to 32.   This shows that different sequences required different number of function evaluations for learning better representations. Training time for an epoch for the models are CDR-NDE-heat(Euler) : 22 sec, CDR-NDE-Heat(Dopri5) : 30 sec, and CDR-NDE : 58 sec, and shows that CDR-NDE-heat models are faster. 

\subsection{Walker2d kinematic simulation} 
The dataset was created by ~\citep{lechner2020learning}  for Walker kinematic modeling task. This is a supervised autoregressive task and the dataset was generated using Walker2d-v2 OpenAI gym environment and MuJoCo physics engine. This dataset evaluates how well a model can simulate kinematic modeling systems that are sampled at irregular time intervals. The training data was generated by performing rollouts on the Walker2d-v2 environment using pre-trained deterministic policy. The Walker environment was trained using a non-recurrent policy though Proximal policy optimization before data collection. The dataset is made irregularly sampled by excluding 10\% of the timesteps. 
The dataset is split into 9684 train, 1937 test, 1272 validation sequences.


In Table \ref{tab:results}, column 3 shows the performance of all the models on Walker2d data. Our proposed model CDR-NDE-heat(Euler and Dopri5) outperform other models with a good margin. The proposed model CDR-NDE also gives a very good performance in this data. Smoothing of the hidden representations allowed the CDR-NDE-heat model to learn well on this data. Again, the more flexible CDR-NDE-heat model using the adaptive Dopri5 solver gives the best performance. Training time for an epoch for the proposed models are CDR-NDE-heat(Euler) : 28 sec, CDR-NDE-Heat(Dopri5) : 48 sec,  and CDR-NDE : 140 sec, and shows that CDR-NDE-heat models are faster. 

 \subsection{Stance Classification}


 In real-world, on social media platforms like twitter, tweets related to a particular event arrive at different times and the inter arrival times between the tweets is not the same. While modeling such a irregular time series data, the hidden representation of each data point could  get affected by their occurrence times. Along with the inter arrival times, one may also need to consider the complexity of an observation or tweet while predicting its class. The complexity level of a tweet or complete sequence of tweets could be of different complexities demanding more complex transformation before making a final prediction. For modeling such a time-series data,  we can learn better representations by considering   continuous transformation of hidden states proportional to the inter arrival times along the axis $t$ and continuous transformation of hidden states depending on the complexity of the sequence along the axis $t'$.  Consequently, our  CDR-NODE models can be a better choice for social media problems such as stance classification.  We evaluated the performance of the models to predict the stance of social media posts~\cite{RumourEval_2019_dataset} under two different experiment setups. This Twitter data set consists of rumours associated with eight events. Each event has collection of tweets labelled with one of the four labels - Support, Query, Deny and Comment. We picked two events, Sydneysiege and Charlihebdo to evaluate the models. Given an event and the corresponding tweets of the event, $N$ number of datapoints each of length 10 are created. The $N$ number of data points will cover the 60\% of the entire tweets of a particular event. While creating a datpoint of a sequence length 10, 10 tweets are randomly selected from the event and then sorting based on the observation time in increasing order. Similarly, the data points for both validation and test data are created by splitting the remaining 40\% equally. 
   \begin{itemize}
       \item \textbf{Seen Event} Here we train, validate and test on tweets of same event. Each
event data is split 60:20:20 ratio in sequence of time. This setup helps in
predicting stance of unseen tweets of the same event.
\item \textbf{Unseen Event} This setup helps in evaluating performance on an unseen event. Here we consider training and validation on
    one event and testing on the other event. 
   \end{itemize}

Tables \ref{tab:sydneysiege} and \ref{tab:Charliehebdo} show the performance of the proposed models compared to the baselines under seen event setup. For the event Sydneysiege as shown in the Table~\ref{tab:sydneysiege}, our proposed models, CDR-NDE-heat(Euler,Dopri5) perform well compared to the baselines in terms of the metrics F1,Recall and Precision. For the event Charliehebdo, as shown in Table~\ref{tab:Charliehebdo} performance of our models CDR-NDE-heat(Euler,Dopri5) is comparable to the performance of other baseline models under the metric F1 and Precision. 
Tables \ref{tab:sydneysiege_unseen} and ~\ref{tab:Charliehebdo_unseen} show the performance of the proposed models compared to the baselines under unseen event setup. For the event Sydneysiege, as shown in the Table~\ref{tab:sydneysiege_unseen}, our proposed models CDR-NDE-heat(Euler,Dopri5) perform well compared to the baselines in terms of the metrics F1 and Precision, and are comparable under the metrics AUC and Recall.  Whereas for the event charliehebdo our models did not beat the baselines in any metric as shown in the Table~\ref{tab:Charliehebdo_unseen}. We observe that in both seen and unseen event setups our models, CDR-NDE-heat(Euler,Dopri5 ) are able to generalize the data well for the event Sydneysiege but are not able to succeed for the event Charliehebdo. 

\begin{table}
        \begin{center}
                \begin{tabular}{|c|c|c|c|c|}
                        \hline
                        {\small \textbf{Model}}  & \multicolumn{4}{c|}{\textbf{Sydneysiege} } \\
                        \hline
                         & \textbf{AUC }& \textbf{F1} & \textbf{Recall} & \textbf{Precision} \\
                         \hline
                        CT-RNN & 0.57 $\pm$ 0.00 & 0.57 $\pm$ 0.00 & 0.70 $\pm$ 0.00 & 0.61 $\pm$ 0.00 \\
                        ODE-RNN & 0.55 $\pm$ 0.01 & 0.55 $\pm$ 0.01 & 0.67 $\pm$ 0.01 & 0.57 $\pm$ 0.01 \\
                        ODE-LSTM & 0.56 $\pm$ 0.01 & 0.56 $\pm$ 0.01 & 0.71 $\pm$ 0.00 & 0.64 $\pm$ 0.01 \\
                        CT-GRU & 0.64 $\pm$ 0.01 & 0.64 $\pm$ 0.01 & 0.72 $\pm$ 0.02 & 0.65 $\pm$ 0.02 \\
                        RNN-Decay & 0.63 $\pm$ 0.01 & 0.63 $\pm$ 0.01 & 0.74 $\pm$ 0.00 & 0.66 $\pm$ 0.00 \\
                        Bidirectional-RNN & 0.62 $\pm$ 0.01 & 0.62 $\pm$ 0.01 & 0.73 $\pm$ 0.01 & 0.66 $\pm$ 0.02 \\
                        GRU-D & 0.64 $\pm$ 0.01 & 0.64 $\pm$ 0.01 & 0.73 $\pm$ 0.01 & 0.65 $\pm$ 0.01 \\
                        Phased-LSTM & 0.61 $\pm$ 0.01 & 0.61 $\pm$ 0.01 & 0.72 $\pm$ 0.01 & 0.62 $\pm$ 0.01 \\
                        GRU-ODE & 0.56 $\pm$ 0.00 & 0.56 $\pm$ 0.00 & 0.70 $\pm$ 0.00 & 0.60 $\pm$ 0.01 \\
                        CT-LSTM & 0.64 $\pm$ 0.01 & 0.64 $\pm$ 0.01 & 0.72 $\pm$ 0.01 & 0.66 $\pm$ 0.01 \\
                        Augmented-LSTM & 0.64 $\pm$ 0.01 & 0.64 $\pm$ 0.01 & 0.73 $\pm$ 0.01 & 0.65 $\pm$ 0.01 \\
                        \hline
                        CDR-NDE & 0.57 $\pm$ 0.01 & 0.62 $\pm$ 0.01 & 0.69 $\pm$ 0.02 & 0.64 $\pm$ 0.05 \\ 
CDR-NDE-heat(Euler) & 0.64 $\pm$ 0.01 & 0.68 $\pm$ 0.01 & 0.73 $\pm$ 0.02 & 0.69 $\pm$ 0.02 \\ 
CDR-NDE-heat(Dopri5) & 0.63 $\pm$ 0.01 & 0.68 $\pm$ 0.01 & 0.73 $\pm$ 0.02 & 0.68 $\pm$ 0.02 \\ \hline 
                \end{tabular}
        \end{center}
        \caption{Performance of the models for sydneysiege(seen-event)}
         \label{tab:sydneysiege}
\end{table}

\begin{table}
        \begin{center}
          \begin{tabular}{|c|c|c|c|c|}
                        \hline
                        {\small \textbf{Model}}  & \multicolumn{4}{c|}{\textbf{Charliehebdo} } \\
                        \hline
                         & \textbf{AUC }& \textbf{F1} & \textbf{Recall} & \textbf{Precision} \\
                         \hline
                        CT-RNN & 0.63 $\pm$ 0.01 & 0.63 $\pm$ 0.01 & 0.72 $\pm$ 0.00 & 0.64 $\pm$ 0.00 \\
                        ODE-RNN & 0.59 $\pm$ 0.02 & 0.59 $\pm$ 0.02 & 0.66 $\pm$ 0.02 & 0.59 $\pm$ 0.03 \\
                        ODE-LSTM & 0.61 $\pm$ 0.01 & 0.61 $\pm$ 0.01 & 0.71 $\pm$ 0.00 & 0.64 $\pm$ 0.01 \\
                        CT-GRU & 0.67 $\pm$ 0.01 & 0.67 $\pm$ 0.02 & 0.75 $\pm$ 0.01 & 0.69 $\pm$ 0.02 \\
                        RNN-Decay & 0.67 $\pm$ 0.02 & 0.67 $\pm$ 0.02 & 0.76 $\pm$ 0.01 & 0.70 $\pm$ 0.01 \\
                        Bidirectional-RNN & 0.67 $\pm$ 0.00 & 0.67 $\pm$ 0.01 & 0.76 $\pm$ 0.00 & 0.68 $\pm$ 0.01 \\
                        GRU-D & 0.69 $\pm$ 0.01 & 0.69 $\pm$ 0.01 & 0.77 $\pm$ 0.01 & 0.70 $\pm$ 0.01 \\
                        Phased-LSTM & 0.64 $\pm$ 0.01 & 0.64 $\pm$ 0.01 & 0.72 $\pm$ 0.01 & 0.66 $\pm$ 0.02 \\
                        GRU-ODE & 0.63 $\pm$ 0.01 & 0.63 $\pm$ 0.01 & 0.72 $\pm$ 0.00 & 0.64 $\pm$ 0.01 \\
                        CT-LSTM & 0.66 $\pm$ 0.04 & 0.66 $\pm$ 0.03 & 0.74 $\pm$ 0.04 & 0.67 $\pm$ 0.04 \\
                        Augmented-LSTM & 0.68 $\pm$ 0.00 & 0.68 $\pm$ 0.01 & 0.77 $\pm$ 0.00 & 0.70 $\pm$ 0.00 \\
                        \hline
                        CDR-NDE & 0.60 $\pm$ 0.01 & 0.62 $\pm$ 0.01 & 0.68 $\pm$ 0.02 & 0.65 $\pm$ 0.02 \\ 
CDR-NDE-heat(Euler) & 0.66 $\pm$ 0.01 & 0.68 $\pm$ 0.01 & 0.73 $\pm$ 0.01 & 0.69 $\pm$ 0.02 \\ 
CDR-NDE-heat(Dopri5) & 0.65 $\pm$ 0.01 & 0.68 $\pm$ 0.01 & 0.73 $\pm$ 0.01 & 0.69 $\pm$ 0.02 \\ \hline 
                \end{tabular}
        \end{center}
        \caption{Performance of the models for charliehebdo(seen-event)}
        \label{tab:Charliehebdo}
\end{table}

\begin{table}
        \begin{center}
               \begin{tabular}{|c|c|c|c|c|}
                        \hline
                        {\small \textbf{Model}}  & \multicolumn{4}{c|}{\textbf{Sydneysiege} } \\
                        \hline
                         & \textbf{AUC }& \textbf{F1} & \textbf{Recall} & \textbf{Precision} \\
                         \hline
                        CT-RNN & 0.56 $\pm$ 0.00 & 0.56 $\pm$ 0.00 & 0.70 $\pm$ 0.00 & 0.61 $\pm$ 0.00 \\
                        ODE-RNN & 0.55 $\pm$ 0.00 & 0.55 $\pm$ 0.00 & 0.67 $\pm$ 0.01 & 0.57 $\pm$ 0.01 \\
                        ODE-LSTM & 0.56 $\pm$ 0.00 & 0.56 $\pm$ 0.00 & 0.70 $\pm$ 0.00 & 0.60 $\pm$ 0.01 \\
                        CT-GRU & 0.63 $\pm$ 0.01 & 0.63 $\pm$ 0.01 & 0.75 $\pm$ 0.01 & 0.65 $\pm$ 0.01 \\
                        RNN-Decay & 0.62 $\pm$ 0.00 & 0.62 $\pm$ 0.00 & 0.74 $\pm$ 0.00 & 0.65 $\pm$ 0.01 \\
                        Bidirectional-RNN & 0.61 $\pm$ 0.00 & 0.61 $\pm$ 0.00 & 0.74 $\pm$ 0.00 & 0.67 $\pm$ 0.02 \\
                        GRU-D & 0.63 $\pm$ 0.00 & 0.63 $\pm$ 0.00 & 0.75 $\pm$ 0.00 & 0.67 $\pm$ 0.03 \\
                        Phased-LSTM & 0.58 $\pm$ 0.01 & 0.58 $\pm$ 0.01 & 0.71 $\pm$ 0.01 & 0.65 $\pm$ 0.05 \\
                        GRU-ODE & 0.56 $\pm$ 0.00 & 0.56 $\pm$ 0.00 & 0.70 $\pm$ 0.00 & 0.60 $\pm$ 0.00 \\
                        CT-LSTM & 0.62 $\pm$ 0.00 & 0.62 $\pm$ 0.01 & 0.75 $\pm$ 0.00 & 0.67 $\pm$ 0.01 \\
                        Augmented-LSTM & 0.63 $\pm$ 0.01 & 0.63 $\pm$ 0.01 & 0.75 $\pm$ 0.00 & 0.69 $\pm$ 0.04 \\
                        \hline
                        CDR-NDE  & 0.57 $\pm$ 0.02 & 0.62 $\pm$ 0.04 & 0.67 $\pm$ 0.06 & 0.66 $\pm$ 0.05 \\ 
CDR-NDE-heat(Euler) & 0.62 $\pm$ 0.01 & 0.68 $\pm$ 0.01 & 0.74 $\pm$ 0.01 & 0.69 $\pm$ 0.02 \\ 
CDR-NDE-heat(Dopri5) & 0.62 $\pm$ 0.01 & 0.68 $\pm$ 0.01 & 0.74 $\pm$ 0.01 & 0.71 $\pm$ 0.02 \\ \hline
                \end{tabular}
        \end{center}
        \caption{Performance of the models for sydneysiege((unseen-event)}
        \label{tab:sydneysiege_unseen}
\end{table}

\begin{table}
        \begin{center}
               \begin{tabular}{|c|c|c|c|c|}
                        \hline
                        {\small \textbf{Model}}  & \multicolumn{4}{c|}{\textbf{Charliehebdo} } \\
                        \hline
                         & \textbf{AUC }& \textbf{F1} & \textbf{Recall} & \textbf{Precision} \\
                         \hline
                        CT-RNN & 0.61 $\pm$ 0.01 & 0.61 $\pm$ 0.01 & 0.70 $\pm$ 0.02 & 0.63 $\pm$ 0.02 \\
                        ODE-RNN & 0.57 $\pm$ 0.03 & 0.57 $\pm$ 0.02 & 0.64 $\pm$ 0.04 & 0.60 $\pm$ 0.02 \\
                        ODE-LSTM & 0.59 $\pm$ 0.00 & 0.59 $\pm$ 0.00 & 0.70 $\pm$ 0.00 & 0.65 $\pm$ 0.00 \\
                        CT-GRU & 0.65 $\pm$ 0.02 & 0.65 $\pm$ 0.02 & 0.74 $\pm$ 0.01 & 0.71 $\pm$ 0.03 \\
                        RNN-Decay & 0.63 $\pm$ 0.01 & 0.63 $\pm$ 0.01 & 0.73 $\pm$ 0.01 & 0.69 $\pm$ 0.01 \\
                        Bidirectional-RNN & 0.64 $\pm$ 0.02 & 0.64 $\pm$ 0.02 & 0.73 $\pm$ 0.01 & 0.70 $\pm$ 0.05 \\
                        GRU-D & 0.65 $\pm$ 0.01 & 0.65 $\pm$ 0.01 & 0.74 $\pm$ 0.00 & 0.71 $\pm$ 0.03 \\
                        Phased-LSTM & 0.60 $\pm$ 0.01 & 0.60 $\pm$ 0.01 & 0.69 $\pm$ 0.01 & 0.60 $\pm$ 0.01 \\
                        GRU-ODE & 0.61 $\pm$ 0.00 & 0.61 $\pm$ 0.00 & 0.71 $\pm$ 0.00 & 0.64 $\pm$ 0.01 \\
                        CT-LSTM & 0.65 $\pm$ 0.01 & 0.65 $\pm$ 0.01 & 0.74 $\pm$ 0.01 & 0.69 $\pm$ 0.03 \\
                        Augmented-LSTM & 0.66 $\pm$ 0.00 & 0.66 $\pm$ 0.01 & 0.75 $\pm$ 0.00 & 0.72 $\pm$ 0.04 \\
                        \hline
                        CDR-NDE & 0.55 $\pm$ 0.01 & 0.53 $\pm$ 0.01 & 0.61 $\pm$ 0.03 & 0.66 $\pm$ 0.02 \\
CDR-NDE-heat(Euler) & 0.58 $\pm$ 0.02 & 0.57 $\pm$ 0.02 & 0.63 $\pm$ 0.02 & 0.59 $\pm$ 0.02 \\ 
CDR-NDE-heat(Dopri5) & 0.57 $\pm$ 0.01 & 0.56 $\pm$ 0.01 & 0.64 $\pm$ 0.00 & 0.59 $\pm$ 0.02 \\ \hline 
                \end{tabular}
        \end{center}
        \caption{Performance of the models for charliehebdo(unseen-event)}
        \label{tab:Charliehebdo_unseen}
\end{table}

\section{Conclusion and Future Work}
We proposed novel continuous depth RNN models based on the framework of  differential equations. The proposed models generalize recurrent NODE models by continuously evolving in both the temporal and depth directions. CDR-NDE models the evolution of hidden states using two separate differential equations. The CDR-ODE-heat model is designed based on the framework of 1D-Heat equation and models the evolution of the hidden states across time and depth. The experimental results on person activity recognition and Walker2d kinematics data showed that our proposed models outperformed the baselines and are very effective on irregularly sampled real-world sequence data.   
Currently,  CDR-NDE models are designed based on the  GRU-cell  transformations. We would like to extend it to other transformations as a future work.
The continuous depth recurrent neural differential equations are very flexible and generic RNN models. They will have widespread application on several complex sequence modeling and time series problems, involving  sequences and  inputs with irregular observation times, and  varying complexities and modalities. 

\bibliography{iclr2023_conference}
\bibliographystyle{iclr2023_conference}


\end{document}